\documentclass{article}

\PassOptionsToPackage{numbers, compress}{natbib}

\usepackage[main, final]{neurips_2026}

\usepackage[utf8]{inputenc} 
\usepackage[T1]{fontenc}    
\usepackage{hyperref}       
\usepackage{url}            
\usepackage{booktabs}       
\usepackage{amsfonts}       
\usepackage{nicefrac}       
\usepackage{microtype}      
\usepackage{xcolor}         
\usepackage{graphicx}
\usepackage{amsmath}
\newtheorem{remark}{Remark}
\usepackage{multirow}
\usepackage{wrapfig}
\usepackage{booktabs}
\usepackage{float}

\title{SpikeSTAG: A Dendritic Compartmental Spiking Graph Network for Multivariate Time-Series Forecasting}

\author{%
  Bang Hu \quad 
  Changze Lv \quad 
  Junyi Wang \quad 
  Mingjie Li \\
  \textbf{Xiaoqing Zheng} \quad 
  \textbf{Fan Zhang}\thanks{Corresponding authors.} \quad 
  \textbf{Wei Cao}\footnotemark[1] \\
  School of Computer Science, Fudan University \\
  \texttt{\{hub24, czlv22, wangjunyi25, mjli24\}@m.fudan.edu.cn} \\
  \texttt{\{zhengxq, ffzhang, caow\}@fudan.edu.cn}
}

\begin{document}

\maketitle

\begin{abstract}
  Spiking Neural Networks (SNNs) offer a distinctive paradigm for temporal modeling through their intrinsic membrane potential dynamics.
However, existing SNN-based forecasting methods focus exclusively on temporal processing, lacking mechanisms to capture spatial dependencies among variables. 
To bridge this gap, we propose SpikeSTAG, a neuromorphic spatiotemporal architecture that integrates graph-based spatial reasoning into spike-driven temporal computation. 
Central to our approach is the Dendritic Graph Module, which reinterprets spectral graph convolutions as biological dendritic integration --- coupling multi-scale spatial aggregation with the temporal dynamics of spiking neurons within a single computational unit.
To adaptively fuse the resulting spatial and temporal representations, we further introduce a Dual-Stream Integration Module that employs competitive gating and coincidence-based amplification, selectively enhancing consistent spatiotemporal patterns while suppressing conflicting noise. 
Extensive experiments demonstrate that SpikeSTAG establishes a new state-of-the-art among SNN-based models and surpasses representative ANN baselines, while reducing theoretical energy consumption by approximately 66.7\% compared to Transformer architectures.
\end{abstract}

\section{Introduction}
Spiking Neural Networks (SNNs), often regarded as the third generation of neural networks, distinguish themselves through event-driven computation, biological plausibility, and energy efficiency~\cite{maass1997networks}.  
A defining feature of SNNs is the membrane potential dynamics of their spiking neurons: the nonlinear integration, thresholding, and reset cycle inherently embeds temporal memory into the network architecture, providing a computational primitive for sequential data that traditional Artificial Neural Networks (ANNs) must engineer explicitly through recurrence or attention.
Leveraging this intrinsic temporal advantage, SNNs have achieved competitive or superior performance across a broad range of tasks, including image classification~\cite{hu2021spiking,fang2021deep,ding2021optimal,zhou2022spikformer,yao2023spike}, visual processing~\cite{meftah2010segmentation,escobar2009action}, and sequential recognition~\cite{jeffares2021spike,chen2023unleashing,dominguez2018deep,fang2020multivariate}. 
More recently, SNNs have been successfully applied to time-series forecasting, where their temporal modeling capability enables performance that matches or surpasses ANNs on established benchmarks~\cite{lv2024efficient,shibo2025ts,wuspikf}.

However, multivariate time-series forecasting demands more than temporal modeling alone. 
Accurate prediction in domains such as energy management and intelligent transportation requires the \emph{simultaneous} capture of temporal dynamics and spatial interactions among variables. 
This need has driven the development of spatiotemporal graph networks that jointly model both dimensions: MTGNN~\cite{wu2020connecting} learns adaptive graph topologies while capturing temporal patterns through dilated convolutions, AGCRN~\cite{bai2020adaptive} combines adaptive graph learning with recurrent architectures, and StemGNN~\cite{cao2020spectral} jointly models spatial and temporal dependencies in the spectral domain. 
These works demonstrate that high-quality spatiotemporal forecasting requires \emph{joint} spatial and temporal modeling --- yet this is precisely the capability that current SNN architectures lack.

Existing SNN-based forecasting methods --- including SeqSNN~\cite{lv2024efficient}, TS-LIF~\cite{shibo2025ts}, and SpikF~\cite{wuspikf} — have progressively refined temporal inductive biases, but model temporal dynamics exclusively with no mechanism for spatial dependencies. 
Meanwhile, SNN-GNN hybrids~\cite{zhu2022spiking,dold2021spike,li2023scaling,sun2024spikegraphormer,zhao2024dynamic} have explored graph-based spatial modeling, but remain confined to classification tasks with decoupled spatiotemporal processing (see Section~\ref{sec:related} for detailed analysis).

In this paper, we bridge this gap by proposing \textbf{SpikeSTAG} (\textbf{Spik}ing \textbf{S}patial-\textbf{T}emporal \textbf{A}daptive \textbf{G}raph Network), a neuromorphic architecture that integrates graph-based spatial reasoning into spike-driven temporal computation for multivariate time-series forecasting.
Central to our design is the Dendritic Graph Module, which enforces scale-segregated spatial processing through a block-diagonal weight structure: each Chebyshev polynomial order is independently transformed by branch-specific weights before integration at the soma, coupling multi-scale spatial aggregation with the temporal dynamics of spiking neurons within a single computational unit. 
To adaptively fuse the resulting spatial and temporal representations, we further introduce a Dual-Stream Integration Module that employs competitive gating and coincidence-based amplification, selectively enhancing consistent spatiotemporal patterns while suppressing conflicting noise.
Extensive experiments on four benchmarks demonstrate that SpikeSTAG establishes a new state-of-the-art among SNN-based models and achieves competitive or superior accuracy against representative ANN baselines, securing the best average placement across datasets, achieving mean ranks of $2.4$ for $R^2$ and $3.3$ for $RSE$ --- while reducing theoretical energy consumption by approximately 66.7\% compared to Transformer architectures. Our main contributions are summarized as follows:
\begin{itemize}
\item We propose a Dendritic Graph Module, which introduces scale-segregated graph convolution into spiking neural networks. 
By enforcing independent per-order weight transformations (a block-diagonal inductive bias), this module couples multi-scale spatial aggregation with spike-driven temporal integration within a single neuron --- providing both measurable accuracy gains and interpretable per-scale spatial decomposition.

\item We introduce a bio-inspired Dual-Stream Integration Module that adaptively arbitrates between spatial and temporal representations through competitive gating and coincidence-based amplification. 
This mechanism selectively enhances high-confidence spatiotemporal patterns while suppressing conflicting noise, enabling robust prediction across both steady-state and disrupted conditions.

\item We conduct extensive evaluations on multivariate time-series forecasting benchmarks. 
The results show that our proposed model sets a new state-of-the-art among SNN-based approaches.
It outperforms representative ANN baselines while maintaining the exceptional energy efficiency inherent to neuromorphic computing.
\end{itemize}

\section{Background}

\subsection{Related Work}
\label{sec:related}

Recent work has successfully applied SNNs to time-series forecasting 
by exploiting their intrinsic temporal modeling advantage. 
SeqSNN~\cite{lv2024efficient} established a spike-driven training 
paradigm; TS-LIF~\cite{shibo2025ts} introduced temporal-segment 
spiking neurons; and SpikF~\cite{wuspikf} leverages Fourier-domain 
spike representations for long-horizon prediction. However, these 
methods model temporal dynamics exclusively, with \emph{no mechanism 
to capture spatial dependencies among variables} --- a critical 
limitation for multivariate forecasting.

A parallel line of research has explored integrating SNNs with GNNs. 
Existing approaches employ Poisson encoding for graph message passing 
through spiking neurons~\cite{zhu2022spiking,dold2021spike}, with 
extensions to dynamic graphs via spike-driven 
attention~\cite{li2023scaling}. More recent work has introduced 
spiking graph transformers~\cite{sun2024spikegraphormer} and reactive 
spiking graph networks~\cite{zhao2024dynamic}. Despite these advances, 
existing SNN-GNN hybrids share three limitations:
\begin{enumerate}
    \item \textbf{Confined to classification.} 
    Current methods target discrete tasks such as node 
    classification~\cite{sun2024spikegraphormer,zhao2024dynamic} or 
    action recognition~\cite{yu2022stsc}, rather than 
    continuous-valued regression.
    
    \item \textbf{Point-neuron aggregation.} 
    Spiking neurons are treated as structureless point processes 
    that apply homogeneous spatial aggregation, discarding 
    multi-scale topological information.
    
    \item \textbf{No spatiotemporal coupling for forecasting.} 
    Spatial graph convolution and temporal spike processing remain 
    decoupled, with no mechanism to jointly model the spatial 
    topology and temporal dynamics required by forecasting tasks.
\end{enumerate}

SpikeSTAG addresses all three limitations by upgrading the 
point-neuron model to a dendritic compartmental neuron via the 
Dendritic Graph Module, and introducing the Dual-Stream Integration 
Module for adaptive spatiotemporal fusion. To our knowledge, 
SpikeSTAG is the first SNN-GNN architecture designed for multivariate time-series forecasting.

\subsection{Problem Formulation}
Let $N$ denote the number of synchronous time-series variables, each represented as a node in a graph, observed over $T$ historical time steps. 
At time step $t$, the system state is $\mathbf{Z}_t = [z_{t,1}, \dots, z_{t,N}]^\top \in \mathbb{R}^N$.
Given a historical sequence $\mathbf{X} = \{\mathbf{Z}_{t-T+1}, \dots, \mathbf{Z}_t\} \in \mathbb{R}^{T \times N}$, we learn a mapping $f(\cdot)$ to forecast the future sequence $\mathbf{Y} = \{\mathbf{Z}_{t+1}, \dots, \mathbf{Z}_{t+L}\} \in \mathbb{R}^{L \times N}$ over the next $L$ time steps.

\subsection{Spiking Neural Networks}

SNNs process information through discrete spike events, governed by 
neuronal dynamics such as the Leaky Integrate-and-Fire (LIF) 
model~\cite{gerstner2002spiking}:
\begin{equation}
    \left\{
    \begin{aligned}
        U[t] &= I[t] + H[t-1], \\
        S[t] &= \Theta\left(U[t] - U_{\mathrm{th}}\right), \\
        H[t] &= \tau \cdot U[t] \cdot \left(1 - S[t]\right) 
               + U_{\mathrm{reset}} \cdot S[t],
    \end{aligned}
    \right.
\end{equation}
where $U[t]$ and $I[t]$ denote the membrane potential and input 
current at time step $t$, $H[t]$ is the hidden state, and $\tau$ 
controls potential decay. A spike $S[t]$ is emitted when $U[t]$ 
exceeds the threshold $U_{\mathrm{th}}$, after which the potential 
resets to $U_{\mathrm{reset}}$. Training employs the arctangent 
surrogate gradient~\cite{fang2021deep} to handle the 
non-differentiability of spike generation.

\subsection{Graph Neural Networks}

Graph Neural Networks capture spatial dependencies via convolutions 
over graph-structured data. Given an undirected graph 
$\mathcal{G} = (\mathcal{V}, \mathcal{E}, A)$ with $N$ nodes and 
adjacency matrix $A \in \mathbb{R}^{N \times N}$, the spectral graph 
convolution is defined via the normalized Laplacian 
$L = I_N - D^{-1/2} A D^{-1/2}$ as:
\begin{equation}
    g_\theta \star x = U \, g_\theta(\Lambda) \, U^\top x,
\end{equation}
where $U$ and $\Lambda$ are the eigenvectors and eigenvalues of $L$, 
respectively. In multivariate time-series forecasting, early methods 
such as STGCN~\cite{yu2018spatio} relied on fixed graph structures, 
while MTGNN~\cite{wu2020connecting} and AGCRN~\cite{bai2020adaptive} 
learn adaptive topologies from data~\cite{wu2019graph,jin2024survey}.
These adaptive graph methods form the spatial modeling foundation upon which SpikeSTAG builds.

\section{Method}

\begin{figure*}[t]
\centering
\includegraphics[width=0.9\linewidth]{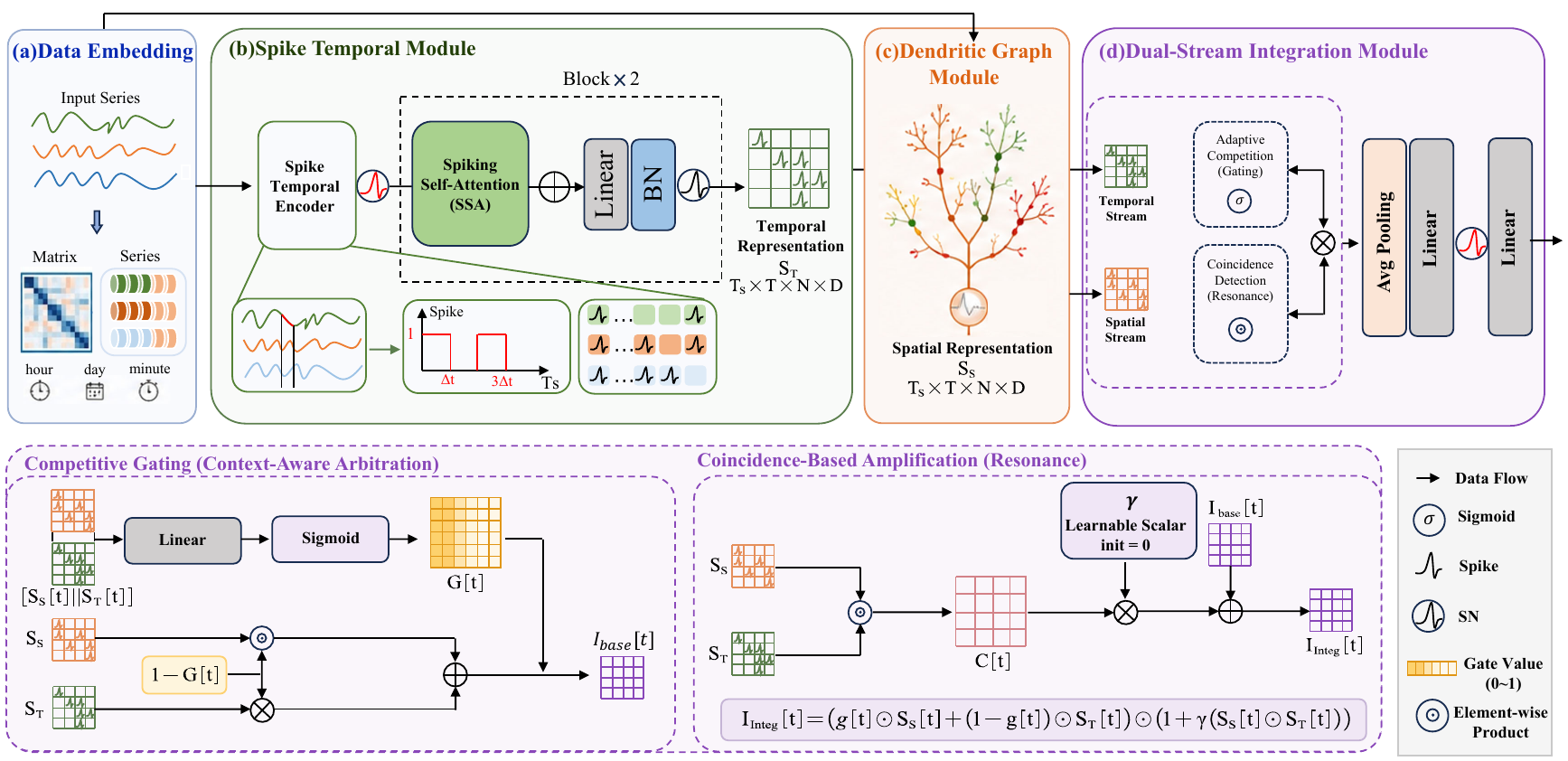}
\caption{Overall architecture of SpikeSTAG. 
(a) Data Embedding: projects inputs and learns an adaptive adjacency matrix. 
(b) Spiking Temporal Module: encodes temporal dynamics via SSA. 
(c) Dendritic Graph Module: extracts multi-scale spatial features. 
(d) Dual-Stream Integration.}
\label{fig2}
\end{figure*}

Figure~\ref{fig2} illustrates the overall architecture of SpikeSTAG.
The embedded representations and adaptive adjacency matrix are fed into two parallel streams:  a \textbf{Dendritic Graph Module} (Section~\ref{sec:dgm}) that extracts multi-scale spatial topologies and a \textbf{Spiking Temporal Module} (Section~\ref{sec:temporal}) that captures temporal dynamics. 
Their outputs are fused by a \textbf{Dual-Stream Integration Module} (Section~\ref{sec:dsim}) to produce the final forecast.

\subsection{Data Embedding}
\label{sec:setup}
Raw node features $\mathbf{Z}_t$ are projected into a high-dimensional space via a linear layer, augmented with learnable temporal embeddings (hour-of-day, day-of-week, and minute-of-hour where applicable) and node identity embeddings:
\begin{equation}
    \mathbf{h}_{t,n}^{0} = \mathrm{Linear}(z_{t,n}) 
    + \mathbf{e}_{n}^{\mathrm{node}} 
    + \mathbf{e}_{h[t]}^{\mathrm{hour}} 
    + \mathbf{e}_{d[t]}^{\mathrm{day}}  
    + \mathbf{e}_{m[t]}^{\mathrm{min}}.
\end{equation}
To model inter-variable spatial dependencies without requiring a predefined graph, we construct an adaptive adjacency matrix from learnable node embeddings $\mathbf{E}$ via $A = \sigma(\mathbf{E}\mathbf{E}^\top) + \lambda \mathbf{I}$, following~\cite{wu2019graph}. 
The self-loop term $\lambda \mathbf{I}$ ensures that each node's own history remains the primary contributor to its future state.

\subsection{Dendritic Graph Module}
\label{sec:dgm}

Existing SNN-GNN hybrids treat spiking neurons as structureless point processes: all spatial scales are aggregated homogeneously before activation, losing the capacity to distinguish local diffusion from global connectivity. We address this by proposing the Dendritic Graph Module, which introduces \emph{scale-segregated} spatial processing into spiking neurons.

\begin{wrapfigure}{R}{0.55\textwidth}
\centering
\vspace{-10pt} 
\includegraphics[width=\linewidth]{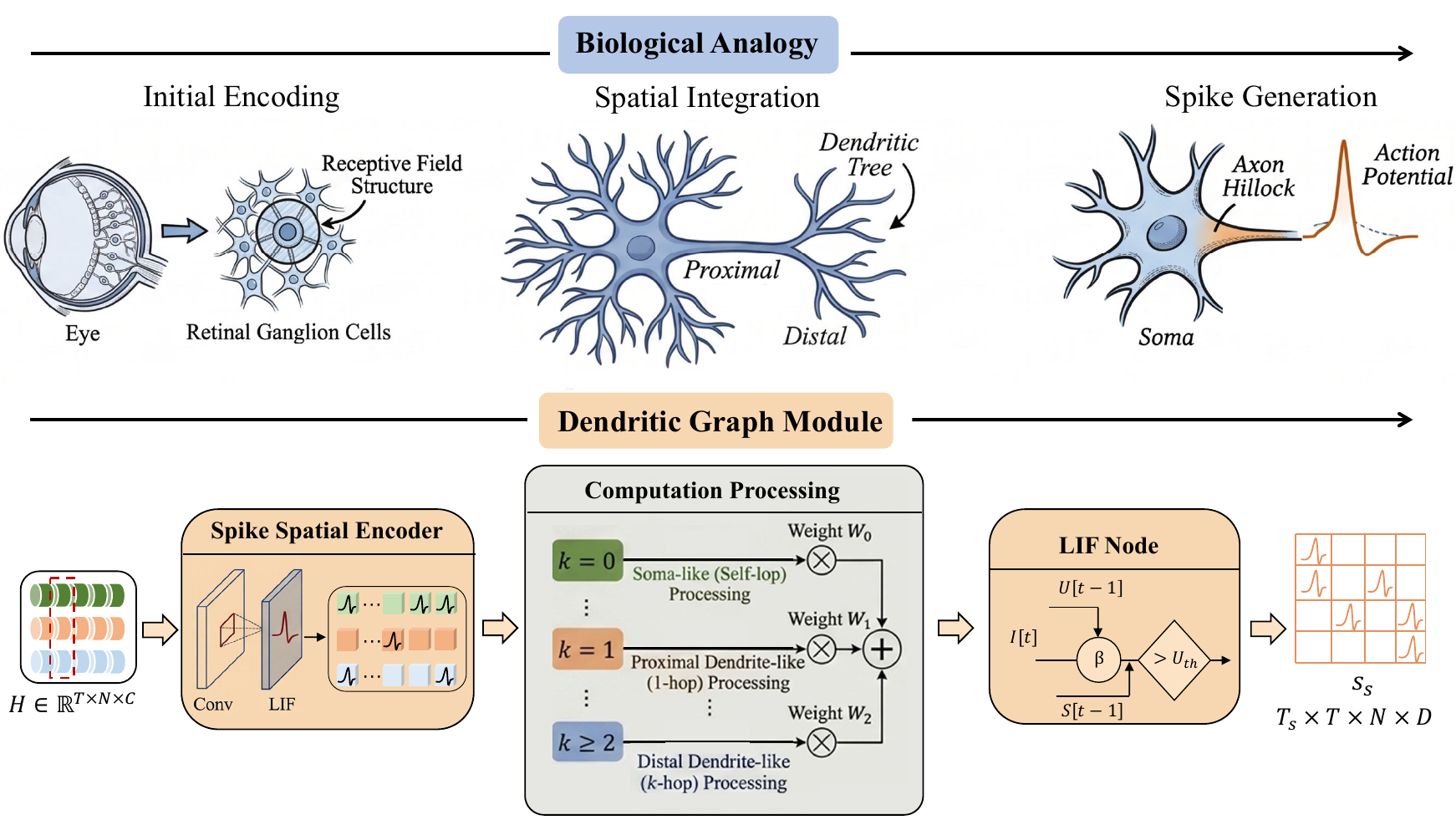}
\caption{Dendritic Graph Module. Top: biological analogy. Bottom: computational model.}
\label{fig3}
\vspace{-10pt} 
\end{wrapfigure}

\subsubsection{Learnable Spike Encoder}

Existing spiking graph networks rely on stochastic Poisson or Bernoulli coding to convert continuous signals into spike trains~\cite{zhu2022spiking,zhao2024dynamic}, introducing uncontrollable noise and precluding gradient-based feature adaptation. We instead employ a deterministic, CNN-based encoder. 
Given continuous input 
$\mathbf{H} \in \mathbb{R}^{T \times N \times C}$:
\begin{equation}
\begin{cases}
    I_{\mathrm{enc}} = \mathrm{Conv1D}(\mathbf{H},\, \mathcal{K}_{\mathrm{enc}}) \\
    S_{\mathrm{in}} = \mathrm{LIF}(I_{\mathrm{enc}})
\end{cases}
\end{equation}
where $S_{\mathrm{in}} \in \{0, 1\}^{T_s \times T \times N \times C'}$ is the binary spike sequence. 
End-to-end backpropagation enables the encoder to learn task-specific spike patterns that preserve spatiotemporal semantics for downstream dendritic integration.

A key consequence of operating on binary spike inputs is the emergence of \emph{dynamic effective graph propagation}: at each simulation step $t$, only active nodes with $S_{\mathrm{in}}[t] = 1$ act as message sources during Chebyshev spatial aggregation, while quiescent nodes are naturally gated out. Formally, this induces an activity-dependent propagation operator $\mathbf{A}_{\mathrm{eff}}^{(k)}[t] = T_k(\tilde{L}) \cdot \mathrm{diag}(S_{\mathrm{in}}[t])$. This formulation realizes an input-dependent, time-varying message-passing schedule where spatial aggregation is dynamically modulated by temporal spike activity across simulation steps without incurring the quadratic query-key computational cost of dense attention.

\subsubsection{Scale-Segregated Dendritic Integration}

\textbf{Spectral Graph Convolution Background.} 
Standard Chebyshev graph convolution approximates localized spectral filters via truncated Chebyshev polynomials:
\begin{equation}
    y = \sigma\!\left(\sum_{k=0}^{K} T_k(\tilde{L})\,\mathbf{X}\, W_k\right),
    \label{eq:chebconv}
\end{equation}
where $W_k \in \mathbb{R}^{C' \times C}$ denotes the order-specific parameter matrix and $\sigma(\cdot)$ is a stateless activation. In conventional GNN-SNN hybrids, spiking neurons treat spatial aggregations as structureless point inputs, passively blending multi-scale representations before reaching the soma and discarding cross-scale temporal dynamics.

\textbf{Dendritic Graph Module (Compartmental Neuron).} 
To bridge this gap, we reinterpret multi-scale Chebyshev operations through the lens of biological dendritic organization.
At each node $n$, the multi-scale spatial inputs from the graph are structured into order-specific dendritic branches with dedicated weights $W_k$, which are then integrated at the soma:
\begin{equation}
\begin{bmatrix}
I_{0,n}[t] \\
\vdots \\
I_{K,n}[t]
\end{bmatrix}
=
\operatorname{blkdiag}(W_0, \dots, W_K) \, z_n[t],
\qquad
I_{\mathrm{total},n}[t] = \sum_{k=0}^{K} I_{k,n}[t],
\label{eq:dgm}
\end{equation}
where $z_n[t] = \big[[T_0(\tilde{L})S_{\mathrm{in}}[t]]_{n,:}, \dots, [T_K(\tilde{L})S_{\mathrm{in}}[t]]_{n,:}\big]^\top \in \mathbb{R}^{(K+1)C}$ concatenates the multi-scale graph signals reaching node $n$, $W_k \in \mathbb{R}^{C' \times C}$, and $I_{k,n}[t] \in \mathbb{R}^{C'}$ is the current generated at dendritic compartment $k$. Here, $t$ indexes SNN simulation steps, with batch and historical sequence dimensions omitted for clarity.

\begin{remark}[Branch-Specific Structural Organization]
\label{remark:block_diag}
The formulation in Eq.~(\ref{eq:dgm}) explicitly organizes scale-isolated linear transformations and preserves intermediate branch outputs prior to somatic summation, facilitating scale-level attribution and intervention. 
While linearly transforming concatenated features followed by unweighted summation is algebraically equivalent to a block-structured projection, this compartmentalized structure provides an interpretable inductive layout where the parameter magnitude $|W_k|$ reflects the relative capacity allocated to each topological scale (cf. Figure~\ref{case}a).
\end{remark}

Drawing a functional analogy with biological dendritic organization, we interpret each polynomial order as a compartment processing signals at a characteristic spatial range:
\begin{itemize}
    \item $k\!=\!0$ (\textit{Soma}): Captures the node's own historical state via self-loop dependencies ($W_0$).
    \item $k\!=\!1$ (\textit{Proximal}): Models local spatial diffusion from immediate 1-hop neighbors ($W_1$), critical for short-range interactions.
    \item $k\!\geq\!2$ (\textit{Distal}): Captures long-range global dependencies via high-order graph skip-connections ($W_k$), filtering distant signals without interference from dominant local patterns.
\end{itemize}

\textbf{Spatiotemporal Coupling.}
The aggregated dendritic current $I_{\mathrm{total}}[t]$ drives the membrane potential dynamics of a LIF neuron:
\begin{equation}
U[t] = \tau U[t-1](1 - S_S[t-1]) + U_{\mathrm{reset}} S_S[t-1] + I_{\mathrm{total}}[t],
\label{eq:membrane}
\end{equation}
where $\tau \in (0, 1)$ is the membrane decay constant. 
Here, the spatial evidence from preceding SNN simulation steps directly modulates the current firing state. 
We emphasize that the DGM adheres to standard LIF neuronal dynamics; its biological analogy serves to structure the multi-scale spatial compartmentalization and somatic integration, rather than formulating a non-standard membrane differential equation. 
This formulation creates a fundamental difference from the standard ``ChebConv $\to$ activation'' pipeline: spatial aggregation at simulation step $t$ is modulated by the residual membrane state from $t\!-\!1$. If $U[t\!-\!1]$ is already near threshold due to accumulated spatial evidence from previous steps, even a sub-threshold dendritic current at time $t$ can trigger a spike. 
This temporal evidence accumulation over spatial signals constitutes the mechanism by which the Dendritic Graph Module achieves joint spatiotemporal encoding within a single computational unit. 
The output spike sequence $S_S$ serves as a spatial structure-aware representation for downstream integration.

\subsection{Spiking Temporal Module}
\label{sec:temporal}
Although the proposed Dendritic Graph Module effectively captures the topological structure of the data, time-series forecasting inherently requires modeling dependencies across multiple time steps.
Consequently, we use a Spiking Temporal Module to address this need.

First, drawing inspiration from the temporal encoder proposed in SeqSNN~\cite{lv2024efficient}, we classify time-series shapes as interpretable primitive features.
We partition a single time step $\Delta T$ of the original series into $T_s$ simulation segments (where $\Delta T = T_s \Delta t$). 
Within each segment, a neuron triggers a firing event if its membrane potential exceeds the threshold. 
This mechanism effectively encodes the embedding layer features into a spike sequence.

Meanwhile, Transformers have gained prominence in time-series forecasting, utilizing multi-head self-attention to effectively model long-range temporal dependencies.
We leverage the Spiking Self-Attention (SSA) mechanism from Spikformer~\cite{zhou2022spikformer} to replace the canonical self-attention layers within the iTransformer architecture~\cite{liu2023itransformer}.
This integration yields a Spiking Transformer block that processes the spike sequences to extract high-level temporal representations $S_T$.

\subsection{Dual-Stream Integration Module}
\label{sec:dsim}

The Spatial stream ($S_S$) and Temporal stream ($S_T$) capture complementary aspects of the data: spatial topology and temporal trends may either diverge (e.g., unexpected accidents disrupting routine patterns) or synchronize (e.g., rush-hour congestion propagating through a sensor network). A robust integration mechanism must handle both regimes --- selectively arbitrating under conflict, while amplifying aligned signals.

Drawing a functional analogy with coincidence detection in cortical pyramidal neurons~\cite{larkum1999new}, we design the Dual-Stream Integration Module as a unified processor that combines competitive gating with resonance-based amplification.

\textbf{Competitive Gating.}
Under divergent conditions, the model must prioritize the more informative stream. We introduce a context-aware gating factor $\mathcal{G}[t] = \sigma(W_g \cdot [S_S[t] \| S_T[t]] + b_g)$, where $\sigma(\cdot)$ is the sigmoid activation and $\|$ denotes concatenation. The base integrated signal is then computed as a competitive mixture:
\begin{equation}
    I_{\text{base}}[t] = 
    \mathcal{G}[t] \odot S_S[t] + 
    (1 - \mathcal{G}[t]) \odot S_T[t],
\end{equation}
where high $\mathcal{G}$ emphasizes spatial topology (e.g., during structural disruptions) and low $\mathcal{G}$ preserves temporal trends (e.g., during steady-state evolution).

\textbf{Coincidence-Based Amplification.}
When both streams are simultaneously active at the same feature dimension, their agreement constitutes high-confidence evidence. 
We detect this via a coincidence factor $\mathcal{C}[t] = S_S[t] \odot S_T[t]$ and modulate the base signal with a learnable gain $\gamma$:
\begin{equation}
    I_{\text{Integ}}[t] = I_{\text{base}}[t] \cdot 
    (1 + \gamma \cdot \mathcal{C}[t]).
\end{equation}
The gain $\gamma$ is initialized to zero, so the model starts from a pure gating regime and progressively learns to amplify resonant patterns during training. This formulation unifies both regimes: the module acts as a stable arbiter under asynchronous conditions and transitions into a sensitive amplifier when spatiotemporal signals resonate.

\textbf{Output Generation.}
The integrated current is temporally aggregated via average pooling to produce a node representation $R = \frac{1}{T_s}\sum_{t=1}^{T_s} I_{\text{Integ}}[t] \in \mathbb{R}^{N \times D}$, which is decoded through a LIF-gated prediction head:
\begin{equation}
    \hat{Y} = \mathrm{Linear}(\mathrm{LIF}(\mathrm{Linear}(R))).
\end{equation}
The prediction head employs LIF neurons to preserve the event-driven nature of SNNs throughout the entire pipeline.

\begin{table*}[htp]
\centering
\caption{Experimental results on 4 benchmark datasets with varying prediction horizons $L$. 
Upper: ANN baselines; Lower: SNN baselines. 
\textbf{Bold}: best; \underline{underline}: second-best. 
$\uparrow$/$\downarrow$: higher/lower is better. 
All results are averages of 3 random seeds (mean$\pm$std reported in Appendix~\ref{app:robustness}).
Baseline sources: ARIMA, GP, iTransformer, iSpikformer, and Autoformer results from~\cite{lv2024efficient}; TS-TCN, TS-GRU, and TS-former from~\cite{shibo2025ts}; 
PatchTST~\cite{nie2023a} implemented and evaluated within the unified framework of~\cite{lv2024efficient}.}
\label{tab:full_results}
\setlength{\tabcolsep}{5pt}
\resizebox{\textwidth}{!}{
    \begin{tabular}{@{}l l *{16}{c} c c @{}}
    \toprule
    \multirow{2.5}{*}{Method} & \multirow{2.5}{*}{Metric} & \multicolumn{4}{c}{METR-LA} & \multicolumn{4}{c}{PEMS-BAY} & \multicolumn{4}{c}{Solar} & \multicolumn{4}{c}{Electricity} & \multirow{2.5}{*}{Avg} & \multirow{2.5}{*}{\shortstack{Avg\\Rank$\downarrow$}}\\
    \cmidrule(lr){3-6} \cmidrule(lr){7-10} \cmidrule(lr){11-14} \cmidrule(lr){15-18}
     & & 6 & 24 & 48 & 96 & 6 & 24 & 48 & 96 & 6 & 24 & 48 & 96 & 6 & 24 & 48 & 96 & & \\ 
    \midrule
    \multirow{2}{*}{ARIMA}
    & $R^2\uparrow$ & $.687$ & $.441$ & $.282$ & $.265$
          & $.741$ & $.723$ & $.692$ & \underline{$.670$} 
          & $.951$ & $.847$ & $.725$ & $.682$ 
          & $.963$ & $.960$ & $.914$ & $.863$ & $.713$ & $8.0$\\
    & $RSE$$\downarrow$  & $.575$ & $.742$ & $.889$ & $.902$
          & $.532$ & $.548$ & $.562$ & $.612$ 
          & $.202$ & $.365$ & $.588$ & $.589$ 
          & $.522$ & $.534$ & $.564$ & $.599$ & $.583$ & $8.1$\\
    \midrule
    \multirow{2}{*}{GP}
    & $R^2\uparrow$ & $.685$ & $.437$ & $.265$ & $.233$ 
          & $.732$ & $.712$ & $.689$ & $.665$ 
          & $.944$ & $.836$ & $.711$ & $.675$ 
          & $.962$ & $.968$ & $.912$ & $.852$ & $.705$ & $9.3$\\
    & $RSE$$\downarrow$  & $.572$ & $.738$ & $.912$ & $.925$ 
          & $.544$ & $.532$ & $.577$ & $.592$
          & $.225$ & $.388$ & $.612$ & $.575$ 
          & $.603$ & $.612$ & $.633$ & $.642$ & $.605$ & $8.7$\\
    \midrule
    \multirow{2}{*}{Autoformer}
    & $R^2\uparrow$ & $.762$ & $.548$ & $.411$ & $.282$ 
          & $.782$ & $.711$ & $.689$ & $.668$ 
          & $.960$ & $.852$ & $.791$ & $.701$ 
          & $.980$ & $.977$ & $.975$ & $.963$ & $.753$ & $6.8$\\
    & $RSE$$\downarrow$  & $.565$ & $.692$ & $.785$ & $.872$ 
          & $.452$ & $.543$ & $.577$ & $.565$
          & $.212$ & $.432$ & $.622$ & $.685$ 
          & $.481$ & $.506$ & $.566$ & $.548$ & $.569$ & $7.3$\\
    \midrule
    \multirow{2}{*}{PatchTST}
    & $R^2\uparrow$ & $.819$ & $.618$ & $.434$ & $.284$ 
          & $.873$ & $.718$ & $.686$ & $.612$ 
          & $.962$ & $.868$ & $.795$ & $.729$ 
          & $.980$ & $.977$ & $.975$ & $.963$ & $.768$ & $5.8$\\
    & $RSE$$\downarrow$  & $.456$ & $.666$ & $.792$ & $.875$ 
          & $.385$ & $.532$ & $.569$ & $.577$
          & $.201$ & $.369$ & $.467$ & $.588$ 
          & $.264$ & $.325$ & $.326$ & $.441$ & $.490$ & $5.7$\\
    \midrule
    \multirow{2}{*}{iTransformer}
    & $R^2\uparrow$ & $.829$ & $\textbf{.623}$ & $.439$ & $.285$
          & $.887$ & $.719$ & $.685$ & $.668$ 
          & $\textbf{.964}$ & $.879$ & $.799$ & $.738$ 
          & $.979$ & $.977$ & $.975$ & $.964$ & $.776$ & $4.3$\\
    & $RSE$$\downarrow$  & $.436$ & $\textbf{.648}$ & $.780$ & $.878$
          & $.362$ & $.547$ & $\textbf{.561}$ & $.584$ 
          & \underline{$.191$} & $.348$ & $.448$ & $.563$ 
          & $.259$ & $.305$ & $.335$ & $.427$ & $.480$ & $4.4$\\
    \midrule
    \midrule
    \multirow{2}{*}{TS-TCN}
    & $R^2\uparrow$ & $.810$ & $.605$ & $\textbf{.473}$ & \underline{$.328$}
          & \underline{$.897$} & \underline{$.759$} & \underline{$.698$} & $.652$ 
          & $\textbf{.964}$ & \underline{$.884$} & $.762$ & $.720$ 
          & $.980$ & $.971$ & $.968$ & $.962$ & $\underline{.777}$ & $4.9$\\
    & $RSE$$\downarrow$  & $.459$ & $.656$ & $.757$ & \underline{$.857$}
          & \underline{$.354$} & \underline{$.527$} & $.590$ & $.633$ 
          & $\textbf{.189}$ & $\textbf{.325}$ & $.484$ & $.523$ 
          & $.264$ & $.316$ & $.318$ & $.360$ & $.476$ & $4.5$\\
    \midrule
    \multirow{2}{*}{TS-GRU}
    & $R^2\uparrow$ & $\textbf{.848}$ & $.618$ & $.430$ & $\textbf{.329}$ 
          & $.874$ & $.742$ & $.684$ & $.649$ 
          & $.938$ & $.878$ & \underline{$.779$} & $.722$ 
          & \underline{$.991$} & $.981$ & \underline{$.983$} & $.976$ & $.776$ & $4.9$\\
    & $RSE$$\downarrow$  & $\textbf{.412}$ & \underline{$.651$} & $.795$ & $\textbf{.853}$ 
          & $.384$ & $.530$ & $.587$ & $.637$
          & $.253$ & $.349$ & \underline{$.426$} & $.527$ 
          & $.216$ & $.240$ & $.236$ & $.271$ & $.460$ & $4.7$\\
    \midrule
    \multirow{2}{*}{iSpikformer}
    & $R^2\uparrow$ & $.817$ & $.618$ & $.440$ & $.279$ 
          & $.879$ & $.744$ & $.687$ & $\textbf{.674}$
          & $.961$ & $.876$ & $.795$ & $.738$ 
          & $.977$ & $.974$ & $.972$ & $.963$ & $.775$ & $5.2$\\
    & $RSE$$\downarrow$  & $.475$ & $.668$ & \underline{$.752$} & $.905$ 
          & $.376$ & $.536$  & \underline{$.569$} & $\textbf{.580}$ 
          & $.204$ & $.333$ & $.465$ & $.521$ 
          & $.263$ & $.284$ & $.338$ & $.348$ & $.476$ & $4.9$\\
    \midrule
    \multirow{2}{*}{TS-former}
    & $R^2\uparrow$ & \underline{$.847$} & \underline{$.620$} & $.445$ & $.283$ 
          & $.874$ & $.735$ & $.683$ & $.669$
          & $.961$ & $\textbf{.886}$ & $\textbf{.828}$ & $\textbf{.774}$
          & $.987$ & \underline{$.985$} & $.981$ & \underline{$.977$} & $.783$ & \underline{$3.4$}\\
    & $RSE$$\downarrow$  & \underline{$.416$} & $.655$ & $.783$ & $.874$ 
          & $.379$ & $.539$ & $.572$ & \underline{$.583$} 
          & $.224$ & $.331$ & $\textbf{.382}$ & $\textbf{.435}$
          & \underline{$.197$} & \underline{$.215$} & \underline{$.234$} & \underline{$.261$} & $\textbf{.441}$ & \underline{$3.4$}\\
    \midrule
    \multirow{2}{*}{SpikeSTAG}
    & $R^2\uparrow$ & $.843$ & $.607$ & \underline{$.468$} & $.270$ 
          & $\textbf{.899}$ & $\textbf{.763}$ & $\textbf{.712}$ & \underline{$.670$} 
          & $.962$ & $.881$ & $.805$ & \underline{$.754$} 
          & $\textbf{.993}$ & $\textbf{.988}$ & $\textbf{.985}$ & $\textbf{.979}$ & $\textbf{.786}$ & $\textbf{2.4}$\\
    & $RSE$$\downarrow$  & $.418$ & $.661$ & \textbf{.712} & $.895$
          & $\textbf{.343}$ & $\textbf{.518}$ & $.580$ & $.591$ 
          & $.225$ & \underline{$.329$} & $.451$ & \underline{$.510$}
          & $\textbf{.185}$ & $\textbf{.194}$ & \textbf{$.215$} & \textbf{$.260$} & \underline{$.443$} & $\textbf{3.3}$\\
    \bottomrule
    \end{tabular}
}
\end{table*}

\begin{table*}[t]
\centering
\caption{Comparison with MTGNN~\cite{wu2020connecting}, a representative GNN-based forecaster. Both models were evaluated within the unified framework of~\cite{lv2024efficient}. 
\textbf{Bold}: best. All results are averages of 3 random seeds.}
\setlength{\tabcolsep}{2.5pt} 
\renewcommand{\arraystretch}{1.1} 
\resizebox{\linewidth}{!}{
\begin{tabular}{@{}ll *{5}{c} *{5}{c} *{5}{c} *{5}{c} c c@{}}
\toprule
\multirow{2.5}{*}{Method} & \multirow{2.5}{*}{Metric} & \multicolumn{5}{c}{METR-LA} & \multicolumn{5}{c}{PEMS-BAY} & \multicolumn{5}{c}{Solar} & \multicolumn{5}{c}{Electricity} & \multirow{2.5}{*}{Avg} & \multirow{2.5}{*}{\shortstack{Avg\\Rank$\downarrow$}}\\
\cmidrule(lr){3-7} \cmidrule(lr){8-12} \cmidrule(lr){13-17} \cmidrule(lr){18-22}
 & & 3 & 6 & 12 & 24 & 48 & 3 & 6 & 12 & 24 & 48 & 3 & 6 & 12 & 24 & 48 & 3 & 6 & 12 & 24 & 48 & & \\
\midrule

\multirow{2}{*}{SpikeSTAG}
& $R^2\uparrow$
& $\textbf{.892}$ & $\textbf{.843}$ & $\textbf{.751}$ & $\textbf{.607}$ & $\textbf{.468}$
& $\textbf{.931}$ & $\textbf{.899}$ & $\textbf{.822}$ & $\textbf{.768}$ & $\textbf{.712}$
& $\textbf{.965}$ & $\textbf{.962}$ & $\textbf{.930}$ & $\textbf{.881}$ & $\textbf{.805}$
& $\textbf{.994}$ & $\textbf{.993}$ & $.990$ & $.988$ & $.985$
& $\textbf{.859}$ & $\textbf{1.2}$ \\

& $RSE\downarrow$
& $\textbf{.346}$ & $\textbf{.418}$ & $\textbf{.526}$ & $\textbf{.661}$ & $\textbf{.712}$
& $\textbf{.284}$ & $\textbf{.343}$ & $\textbf{.456}$ & $\textbf{.521}$ & $\textbf{.580}$
& $\textbf{.192}$ & $\textbf{.225}$ & $\textbf{.272}$ & $\textbf{.338}$ & $\textbf{.451}$
& $.168$ & $.185$ & $.190$ & $.194$ & $.215$
& $\textbf{.364}$ & $\textbf{1.3}$ \\
\midrule

\multirow{2}{*}{MTGNN}
& $R^2\uparrow$
& $.850$ & $.788$ & $.694$ & $.573$ & $.440$
& $.918$ & $.867$ & $.807$ & $.741$ & $.684$
& $.963$ & $.947$ & $.921$ & $.870$ & $.802$
& $\textbf{.994}$ & $.992$ & $\textbf{.991}$ & $\textbf{.989}$ & $\textbf{.986}$
& $.841$ & $1.8$ \\

& $RSE\downarrow$
& $.407$ & $.484$ & $.582$ & $.687$ & $.787$
& $.298$ & $.381$ & $.459$ & $.532$ & $.587$
& $.198$ & $.237$ & $.288$ & $.370$ & $.457$
& $\textbf{.138}$ & $\textbf{.159}$ & $\textbf{.166}$ & $\textbf{.187}$ & $\textbf{.210}$
& $.381$ & $1.8$ \\

\bottomrule
\end{tabular}%
}
\label{tab:mtgnn_results}
\end{table*}

\section{Experiments}
We evaluate SpikeSTAG through five aspects:
(1) predictive performance across four spatiotemporal benchmarks (Section~\ref{sec:main_results});
(2) component contributions via ablation studies on the Dendritic Graph Module and Dual-Stream Integration Module (Section~\ref{sec:ablation});
(3) interpretability of spatial modeling through a case study (Section~\ref{sec:case_study});
(4) cross-domain adaptability via a generalization study (Section~\ref{sec:generalization});
and (5) computational efficiency via energy consumption analysis (Section~\ref{sec:efficiency}).
Dataset descriptions, evaluation metrics, and implementation details are provided in Appendix~\ref{app:implementation}. 
Robustness analysis across multiple random seeds ($\text{mean}\pm\text{std}$) is reported in Appendix~\ref{app:robustness}.

\subsection{Main Results}
\label{sec:main_results}

\subsubsection{Datasets and Metrics}
We evaluate on two long-sequence (Electricity, Solar) and two short-sequence (METR-LA, PEMS-BAY) datasets, covering energy and traffic domains with diverse temporal granularities.
Following the established SNN forecasting protocol~\cite{lv2024efficient,shibo2025ts}, we adopt the root relative squared error ($RSE$) and the coefficient of determination ($R^2$) as primary metrics due to their scale invariance --- critical for fair comparison across datasets with different numerical magnitudes.

\subsubsection{Performance Comparison}
Table \ref{tab:full_results} presents a comprehensive performance comparison across four benchmark datasets.   
SpikeSTAG achieves the best average rank among all methods ($R^2$: $2.4$, $RSE$: $3.3$), confirming that the proposed architecture delivers consistently strong spatiotemporal modeling across diverse data regimes.

\textbf{Establishing a New SOTA for SNNs:}
SpikeSTAG consistently outperforms all existing SNN variants. 
Pure temporal SNN models (TS-TCN, TS-GRU, TS-former, iSpikformer) exhibit strong competitiveness on traffic datasets where distinct periodicity patterns favor temporal modeling, but their performance degrades on the Electricity dataset where complex latent inter-client dependencies require spatial reasoning. 
By integrating the Dendritic Graph Module, SpikeSTAG overcomes this spatial bottleneck: on Electricity, it achieves $R^2 = 0.993$ at horizon 6 and $0.979$ at horizon 96, substantially outperforming all temporal-only SNN baselines.

\textbf{Surpassing Competitive ANN Baselines:} Remarkably, SpikeSTAG not only dominates the neuromorphic domain but also rivals or surpasses representative ANN models.
On PEMS-BAY (horizon 6), it outperforms iTransformer ($R^2$: 0.899 vs.\ 0.887) and PatchTST ($R^2$: 0.899 vs.\ 0.873). 
On Electricity, SpikeSTAG achieves the highest $R^2$ and lowest $RSE$ across all horizons, demonstrating that spatial graph reasoning combined with spike-driven temporal processing can match the representational capacity of Transformer architectures.

\textbf{Evaluation of Spatial Modeling Capability:} 
To isolate the contribution of spatial modeling, we compare directly against MTGNN~\cite{wu2020connecting}, a representative graph neural network, in Table~\ref{tab:mtgnn_results}. 
SpikeSTAG achieves a superior average rank (1.2 vs.\ 1.8). While results on Electricity are comparable --- validating that both architectures benefit from graph learning for latent correlations --- SpikeSTAG dominates on METR-LA, PEMS-BAY, and Solar across nearly all horizons. 
For instance, on METR-LA (horizon 48), SpikeSTAG achieves $R^2 = 0.468$ vs.\ MTGNN's $0.440$. 
This gap underscores that for systems with rapid temporal evolution, the Dual-Stream Integration Module captures temporal dynamics more effectively than the decoupled temporal processing in MTGNN.

\subsection{Ablation Study}
\label{sec:ablation}
\begin{wrapfigure}{R}{0.55\textwidth}
\centering
\vspace{-10pt} 
\includegraphics[width=\linewidth]{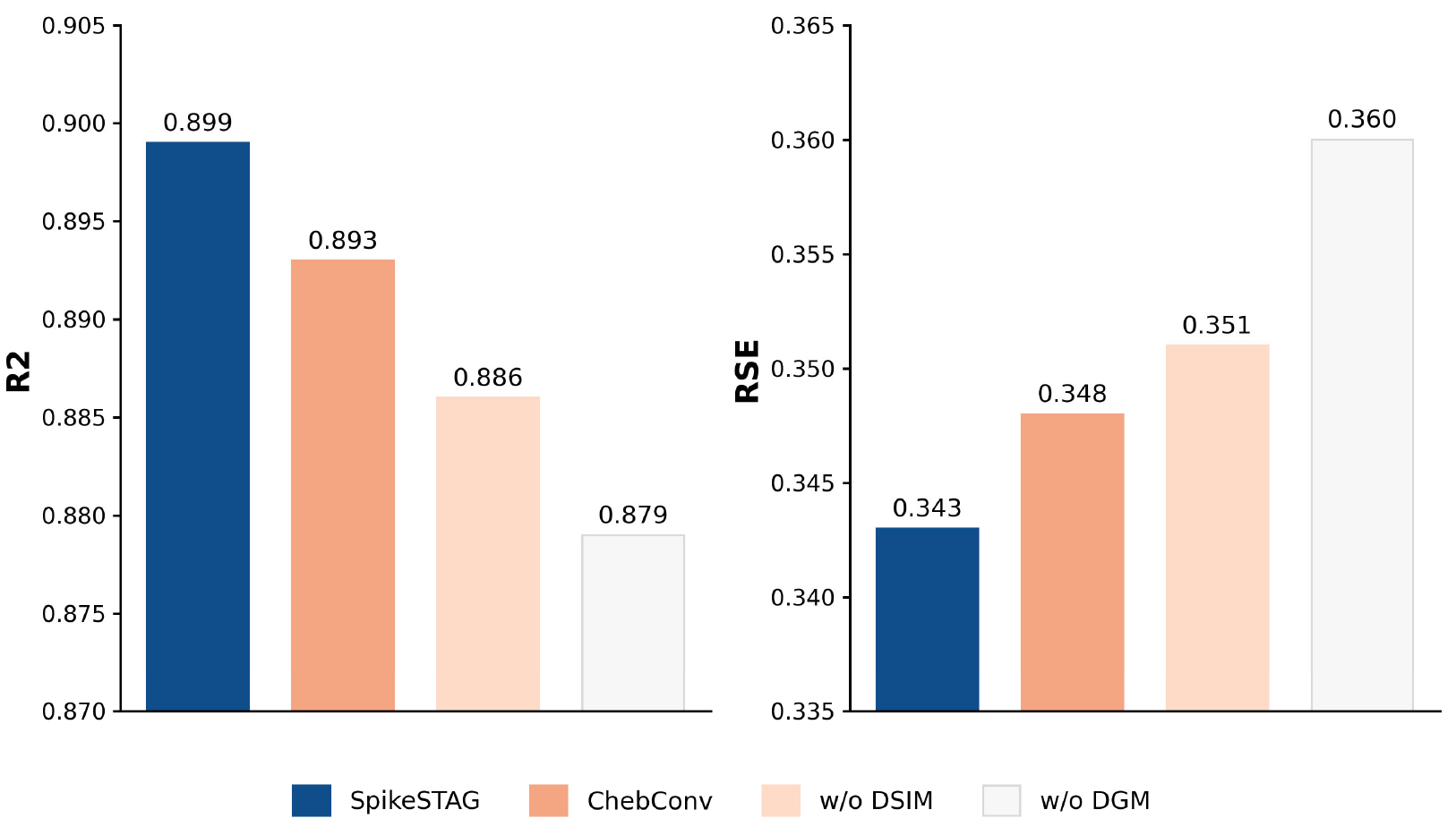}
\caption{Ablation study on PEMS-BAY (Horizon 6).}
\label{abi}
\vspace{-10pt} 
\end{wrapfigure}

To disentangle the specific contributions of each component within our framework, we conducted an ablation study on the PEMS-BAY dataset (horizon 6). 
As illustrated in Figure \ref{abi}, the results reveal a clear performance hierarchy: SpikeSTAG $>$ ChebConv $>$ w/o DSIM $>$ w/o DGM.

\begin{wrapfigure}{R}{0.55\textwidth}
\centering
\vspace{-10pt}
\includegraphics[width=\linewidth]{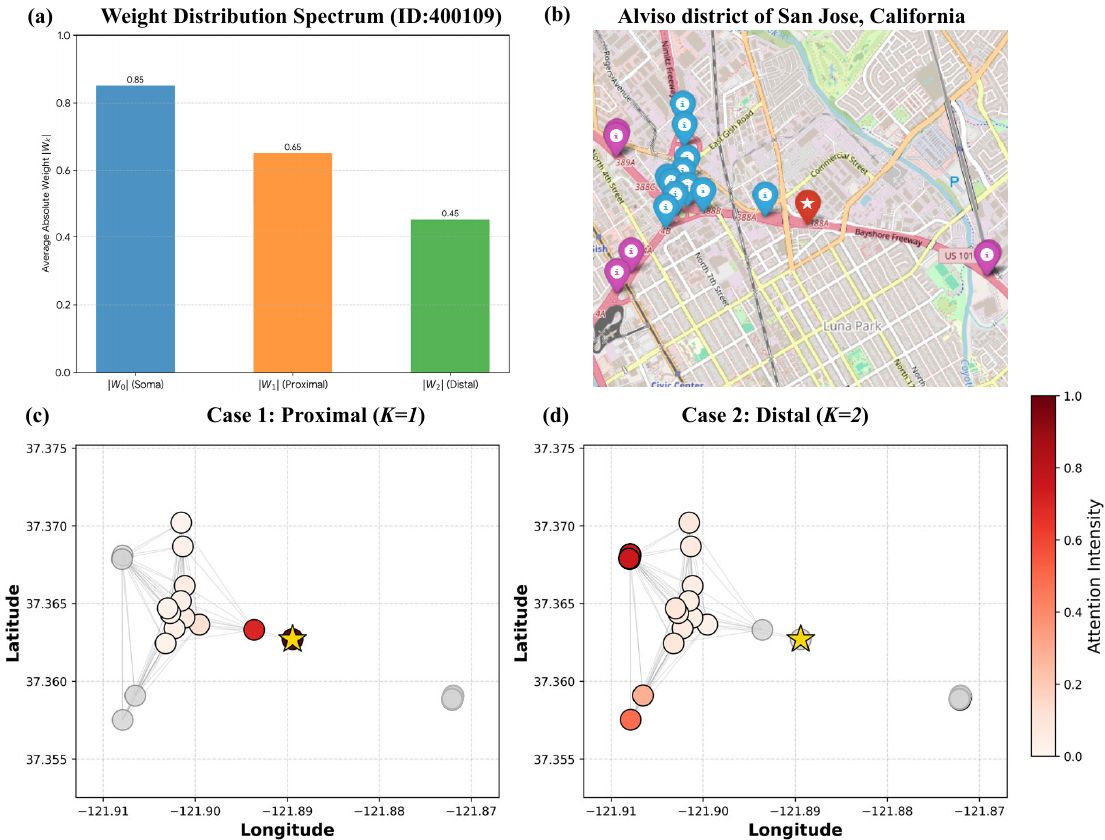}
\caption{Case study on Station 400109, PEMS-BAY. 
(a) Learned weight spectrum $|W_k|$ by spatial order. 
(b) Geographic context. 
(c) Proximal ($K\!=\!1$): local diffusion. 
(d) Distal ($K\!=\!2$): skip-connection to distant nodes.
}
\label{case}
\vspace{-10pt}
\end{wrapfigure}

\textbf{Impact of Spatial Modeling:}
Removing the Dendritic Graph Module (w/o DGM) causes the most significant degradation, confirming that spatial contextualization via scale-segregated dendritic integration is the pivotal factor for state-of-the-art accuracy.
To further verify that this gain stems from the architectural design rather than simply adding graph convolution, we compare against a ``ChebConv$\to$LIF'' baseline (standard dense-weight ChebConv followed by LIF neurons) with detailed results in Appendix~\ref{app:ablation}.

\textbf{Necessity of Dual-Stream Integration:}
Replacing the Dual-Stream Integration Module with a standard residual connection (w/o DSIM) yields intermediate performance, situated between the full model and the w/o DGM variant. 
This confirms that spatial and temporal features are not merely additive: the competitive gating and coincidence-based amplification provide context-aware arbitration that resolves spatiotemporal conflicts and amplifies synchronized patterns.

\newpage 
\subsection{Case Study}
\label{sec:case_study}
In this section, we analyze the Dendritic Graph Module to verify its capacity to 
simultaneously attend to local and global signals. 
We examine a target node (Station 400109) in the Alviso district of San Jose, CA (Figure~\ref{case}b), during the morning peak (08:00 AM).

As shown in Figure~\ref{case}(a), the learned synaptic weights exhibit a decaying yet heavy-tailed distribution: 
the Soma dominates ($|W_0| = 0.85$), reflecting strong autoregressive dependency, while the Distal compartment retains substantial efficacy ($|W_2| = 0.45$). 
This confirms that the model actively leverages high-order connections for long-range dependencies rather than relying solely on local diffusion.
 
Figures~\ref{case}(c) and~\ref{case}(d) visualize the spatial segregation at orders $K\!=\!1$ and $K\!=\!2$. 
For $K\!=\!2$, we highlight the differential contribution relative to the proximal baseline. 
The module displays a distinct skip-connection property: it filters out dominant local signals (captured by $K\!=\!1$) to selectively attend to distant peripheral nodes. 
This mechanism enables the Dendritic Graph Module to anticipate traffic waves propagating from upstream or downstream arterial roads, transcending the immediate local congestion horizon.
This per-scale decomposition --- directly readable from the branch-specific weights $W_k$ --- is a qualitative property enabled by the block-diagonal inductive bias.

\subsection{Generalization Study}
\label{sec:generalization}

To evaluate whether SpikeSTAG generalizes beyond physical spatial graphs, we test its transferability on three canonical LTSF benchmarks lacking predefined topologies: \textbf{Weather}, \textbf{Exchange}, and \textbf{ETTh1}. 
We compare against the representative temporal SNN baseline iSpikformer~\cite{zhou2022spikformer} and the ANN baseline iTransformer~\cite{liu2023itransformer} across forecasting horizons $L \in \{6, 24, 48, 96\}$.

\begin{table*}[h]
\centering
\small
\setlength{\tabcolsep}{3.5pt}
\caption{Zero-prior topology generalization on LTSF benchmarks. Each cell reports $R^2$ / $RSE$ across varying horizons $L$. The comparison validates inductive transferability against representative temporal SNN and ANN architectures rather than an exhaustive baseline ranking.}
\label{tab:ltsfresults}
\resizebox{\textwidth}{!}{%
\begin{tabular}{l c cccc cccc cccc}
\toprule
\multirow{2}{*}{\textbf{Method}} & \multirow{2}{*}{\textbf{Metric}} & \multicolumn{4}{c}{\textbf{Weather}} & \multicolumn{4}{c}{\textbf{Exchange}} & \multicolumn{4}{c}{\textbf{ETTh1}} \\
\cmidrule(lr){3-6} \cmidrule(lr){7-10} \cmidrule(lr){11-14}
& & H=6 & H=24 & H=48 & H=96 & H=6 & H=24 & H=48 & H=96 & H=6 & H=24 & H=48 & H=96 \\
\midrule
SpikeSTAG & $R^2\uparrow$ & \textbf{0.983} & \textbf{0.972} & \textbf{0.965} & \textbf{0.957} & \textbf{0.9997} & \textbf{0.9988} & \textbf{0.9979} & \textbf{0.9967} & \textbf{0.8350} & \textbf{0.7419} & \textbf{0.6907} & 0.6309 \\
          & $RSE\downarrow$ & \textbf{0.450} & \textbf{0.578} & \textbf{0.654} & \textbf{0.723} & \textbf{0.1242} & \textbf{0.2140} & \textbf{0.2873} & \textbf{0.3916} & \textbf{0.4490} & \textbf{0.5606} & \textbf{0.6132} & 0.6681 \\
\midrule
iSpikformer & $R^2\uparrow$ & 0.980 & 0.965 & 0.958 & 0.950 & 0.9990 & 0.9981 & 0.9971 & 0.9951 & 0.7526 & 0.6060 & 0.5954 & 0.5798 \\
            & $RSE\downarrow$ & 0.499 & 0.620 & 0.690 & 0.756 & 0.1918 & 0.2699 & 0.3377 & 0.4458 & 0.5488 & 0.6927 & 0.7019 & 0.7150 \\
\midrule
iTransformer & $R^2\uparrow$ & 0.980 & 0.971 & 0.964 & 0.956 & 0.9996 & \textbf{0.9988} & 0.9978 & 0.9958 & 0.8085 & 0.7357 & 0.6825 & \textbf{0.6406} \\
             & $RSE\downarrow$ & 0.467 & 0.594 & 0.664 & 0.734 & 0.1250 & 0.2198 & 0.2946 & 0.4161 & 0.4829 & 0.5674 & 0.6218 & \textbf{0.6613} \\
\bottomrule
\end{tabular}%
}
\end{table*}

As shown in Table~\ref{tab:ltsfresults}, SpikeSTAG consistently outperforms iSpikformer (e.g., boosting $R^2$ from 0.7526 to 0.8350 on ETTh1 at $H=6$) and rivals or exceeds iTransformer across horizons. 
This transferability stems from two complementary architectural properties: 
(1) mathematically, Chebyshev graph convolutions act as scale-segregated spectral filters independent of physical coordinates; coupled with the adaptive adjacency $A = \sigma(E E^\top) + \lambda I$, DGM discovers latent semantic manifolds while isolating self-dynamics ($k=0$) from cross-variable interference ($k \ge 1$); and 
(2) event-driven spike sparsity naturally suppresses quiescent channels, preventing spurious correlation noise across arbitrary multivariate variables.

\subsection{Efficiency Analysis}
\label{sec:efficiency}

SpikeSTAG preserves energy efficiency despite its dual-stream architecture. 
Following the established estimation protocol based on 45nm CMOS technology~\cite{horowitz20141} (MAC $= 4.6$ pJ, AC $= 0.9$ pJ), we report energy consumption of the encoder backbone excluding task-specific prediction heads. 
All spike rates were empirically measured over 200 inference batches.

\begin{wraptable}{R}{0.55\textwidth}
    \centering
    \small 
    \vspace{-10pt} 
    \caption{Theoretic energy consumption per sample of PEMS-BAY (Inference Stage).}
    \label{Energy}
    \setlength{\tabcolsep}{4pt} 
    \begin{tabular}{l c c c c c}
        \toprule
        \textbf{Model} & \textbf{Param} & \textbf{OPs} & \textbf{Spike} & \textbf{Energy} & \textbf{$R^2$} \\
        & \textbf{(M)} & \textbf{(G)} & \textbf{Rate} & \textbf{(mJ)} & \\
        \midrule
        \textbf{SpikeSTAG}       & $\textbf{0.232}$ & $\textbf{2.106}$ & $\textbf{0.116}$ & $\textbf{2.251}$ & $\textbf{0.899}$ \\ 
    iTransformer    & $1.583$ & $1.470$ & -       & $6.762$ & $0.887$\\ 
    PatchTST        & $0.397$ & $0.782$ & -       & $3.591$ & $0.873$\\ 
    MTGNN           & $0.542$ & $0.256$ & -       & $1.177$ & $0.867$\\
        \bottomrule
    \end{tabular}
    \vspace{-10pt} 
\end{wraptable}
 
As shown in Table~\ref{Energy}, SpikeSTAG achieves 2.25 mJ total energy --- a 66.7\% reduction compared to iTransformer (6.76 mJ) and 37.3\% compared to PatchTST (3.59 mJ) --- while attaining the highest $R^2$. 
This efficiency stems from the SNN core's reliance on AC operations (0.9 pJ each) rather than MACs (4.6 pJ each): although the total operation count is higher, the per-operation energy cost is $5.1\times$ lower.
The Dendritic Graph Module contributes only 0.63 mJ (28\% of SNN energy), validating that scale-segregated Chebyshev message passing incurs substantially lower energy overhead than token-wise self-attention in the SSA layer. Although asymptotic complexity varies with graph sparsity, feature channels, and temporal unrolling, our module-wise energy breakdown (Table~\ref{tab:energy_comparison} in Appendix~\ref{app:energy}) confirms that the sparse, linear-scaling spatial propagation of DGM is substantially more energy-frugal than the quadratic all-to-all interactions in temporal attention under real spike firing rates.
 
We note that MTGNN consumes less total energy (1.18 mJ) due to its compact GCN-only architecture without temporal modeling, but at the cost of lower accuracy ($R^2 = 0.867$ vs.\ $0.899$).
A detailed per-module breakdown with MAC/SOP separation and per-layer spike statistics is provided in Appendix~\ref{app:energy}. 
Preliminary hardware co-design considerations are discussed in Appendix~\ref{app:hardware}.
 
We emphasize that these estimates are theoretical and based on established community protocols~\cite{lv2024efficient,shibo2025ts}. 
Deployment on physical neuromorphic hardware remains an important future direction.

\section{Conclusion}
In this study, we propose SpikeSTAG to bridge the critical gap of spatial representation learning in SNN-based time-series forecasting. 
Our architecture features a Dendritic Compartmental Module that achieves unified encoding of multi-scale spatial topologies within a single neuron, and a Dual-Stream Integration mechanism that adaptively resolves spatiotemporal conflicts and amplifies high-confidence resonance.
Extensive experiments demonstrate that SpikeSTAG establishes a new benchmark for SNNs, outperforming advanced ANN baselines (e.g., iTransformer, MTGNN) while maintaining significantly lower energy consumption relative to Transformer architectures.
Consequently, SpikeSTAG provides a robust, energy-efficient solution for high-performance time-series forecasting and paves the way for applying biologically plausible computing to intricate graph-structured domains.

\section*{Acknowledgments}
The authors would like to thank the anonymous reviewers for their valuable comments.

\section*{Broader Impact}
\label{sec:broader_impact}
This work aims to advance the field of spiking neural networks (SNNs). 
Unlike artificial neural networks (ANNs) which have been applied widely in people’s lives, SNNs are still undergoing fundamental research.
We do not see any negative societal impacts of this work.

\newpage
{
\small
\bibliographystyle{unsrtnat}
\bibliography{example_paper}
}

\newpage
\appendix

\section{Training and Implementation Details}
\label{app:implementation}

\subsection{Datasets Description}

We evaluate on four primary benchmarks --- Solar, Electricity, METR-LA, and PEMS-BAY --- spanning the energy and traffic domains with distinct temporal granularities, together with three supplementary long-term time-series forecasting (LTSF) benchmarks ---Weather, Exchange, and ETTh1 --- that lack predefined spatial topologies and serve to test generalization to pure LTSF settings. \textbf{Solar} records the power production of 137 photovoltaic plants in Alabama in 2006, sampled every 10 minutes, posing significant challenges due to the intermittency of solar energy. \textbf{Electricity} records hourly consumption (kWh) of 321 clients from 2012 to 2014, a long-sequence task with complex seasonal patterns. \textbf{METR-LA} contains traffic speed statistics from 207 loop detectors on the highways of Los Angeles County, a canonical benchmark for dense spatial dependency modeling. \textbf{PEMS-BAY} provides traffic speed data from 325 sensors in the San Francisco Bay Area, emphasizing spatiotemporal interactions over a larger graph topology. \textbf{Weather} records 21 meteorological indicators every 10 minutes throughout 2020 from the Max Planck Biogeochemistry Institute. \textbf{Exchange} contains the daily exchange rates of eight countries from 1990 to 2016, a non-stationary series with weak seasonality. \textbf{ETTh1} records seven indicators of an electricity transformer at hourly granularity from 2016 to 2018. Table~\ref{tab:datasets} summarizes their statistics.

\begin{table}[H]
    \centering
    \small
    \caption{Summary of statistics for benchmark datasets.}
    \label{tab:datasets}
    \begin{tabular}{l l c c c c}
    \toprule
    \textbf{Dataset} & \textbf{Samples} & \textbf{Nodes} & \textbf{Sample Rate} & \textbf{Input Len.} & \textbf{Output Len.} \\ 
    \midrule
    Solar       & 52{,}560 & 137 & 10 min & 168 & $\{6, 24, 48, 96\}$ \\
    Electricity & 26{,}304 & 321 & 1 hour & 168 & $\{6, 24, 48, 96\}$ \\
    METR-LA     & 34{,}272 & 207 & 5 min  & 12  & $\{6, 24, 48, 96\}$ \\
    PEMS-BAY    & 52{,}116 & 325 & 5 min  & 12  & $\{6, 24, 48, 96\}$ \\
    Weather     & 52{,}696 & 21  & 10 min & 96  & $\{6, 24, 48, 96\}$ \\
    Exchange    & 7{,}207  & 8   & 1 day  & 96  & $\{6, 24, 48, 96\}$ \\
    ETTh1       & 17{,}420 & 7   & 1 hour & 96  & $\{6, 24, 48, 96\}$ \\
    \bottomrule
    \end{tabular}
\end{table}

\subsection{Evaluation Metrics}

We employ two primary metrics: the Root Relative Squared Error ($RSE$) and the Coefficient of Determination ($R^2$). These metrics are preferred over MSE or MAE due to their scale invariance, enabling fair comparison across datasets with different numerical magnitudes~\cite{lv2024efficient,shibo2025ts}.

The \textbf{Root Relative Squared Error ($RSE$)} is defined as:
\begin{equation}
\mathrm{RSE} = \sqrt{\frac{\sum_{m=1}^{M} \left\| \mathbf{Y}^{m} - \hat{\mathbf{Y}}^{m} \right\|^{2}}{\sum_{m=1}^{M} \left\| \mathbf{Y}^{m} - \bar{\mathbf{Y}} \right\|^{2}}},
\end{equation}
where values closer to $0$ indicate better performance.

The \textbf{Coefficient of Determination ($R^2$)} is:
\begin{equation}
R^2 = \frac{1}{MCL} \sum_{m=1}^{M} \sum_{c=1}^{C} \sum_{l=1}^{L}
\left[ 1 - \frac{(Y_{c,l}^{m} - \hat{Y}_{c,l}^{m})^2}{(Y_{c,l}^{m} - \bar{Y}_{c,l})^2} \right],
\end{equation}
where $R^2 \in (0, 1]$, with values closer to $1$ indicating superior accuracy. $M$ is the number of test samples, $C$ the number of variables, and $L$ the prediction horizon.

\subsection{Implementation Details}

\textbf{Hardware and Software Environment.} 
All our experiments were conducted on a workstation equipped with an NVIDIA A100 GPU (80GB) and an Intel Core i9-10900K CPU, using PyTorch 1.12 and the SpikingJelly framework.

\textbf{Optimization Strategy and Network Configuration.} 
Models were trained with Adam/AdamW without dropout in spiking layers to preserve spike continuity. Firing threshold is $V_{\mathrm{th}} = 1.0$ with arctangent surrogate gradient ($\alpha = 2.0$), and membrane decay is $\tau = 2.0$ ($\beta = 0.5$). 
Complete training and architectural parameters are summarized in Table~\ref{tab:hyperparams}.

\begin{table}[h]
\centering
\small
\setlength{\tabcolsep}{6pt}
\caption{Hyperparameter configurations of SpikeSTAG across datasets.}
\label{tab:hyperparams}
\begin{tabular}{l c c c c}
\toprule
\textbf{Parameter} & \textbf{Solar} & \textbf{PEMS-BAY} & \textbf{Electricity} & \textbf{METR-LA} \\
\midrule
\multicolumn{5}{l}{\textit{(a) Training Hyperparameters}} \\
Optimizer         & Adam & AdamW & Adam & Adam \\
Learning Rate     & $1\!\times\!10^{-4}$ & $2\!\times\!10^{-3}$ & $1\!\times\!10^{-4}$ & $1\!\times\!10^{-4}$ \\
Weight Decay      & $0.0$ & $0.0$ & $0.0$ & $0.0$ \\
Batch Size        & $4$ & $16$ & $32$ & $4$ \\
Grad. Acc. Steps  & $4$ & $4$ & $1$ & $4$ \\
Early Stop        & $10$ & $20$ & $10$ & $20$ \\
\midrule
\multicolumn{5}{l}{\textit{(b) Architectural Parameters}} \\
Input Dim ($N$)          & $137$ & $325$ & $321$ & $207$ \\
Input Window             & $168$ & $12$ & $168$ & $12$ \\
SNN Hidden Ch.           & $[64,128]$ & $[64,128]$ & $[64,128]$ & $[64,128]$ \\
GCN Hidden Ch.           & $[16,128]$ & $[16,128]$ & $[16,128]$ & $[16,128]$ \\
Chebyshev Order ($K$)    & $2$ & $2$ & $2$ & $2$ \\
Node Embedding Dim       & $16$ & $16$ & $16$ & $16$ \\
\bottomrule
\end{tabular}
\end{table}

\section{Robustness Analysis}
\label{app:robustness}

To verify that performance differences are not artifacts of seed selection, we report SpikeSTAG's results as $\text{mean} \pm \text{std}$ across 3 random seeds on all primary datasets. As shown in Table~\ref{tab:robustness}, standard deviations are consistently $\leq 0.005$ across all dataset-horizon combinations, confirming the architectural robustness of SpikeSTAG.

\begin{table}[h]
\centering
\small
\setlength{\tabcolsep}{4.5pt}
\caption{SpikeSTAG robustness: mean $\pm$ std across 3 random seeds. All standard deviations are $\leq 0.005$.}
\label{tab:robustness}
\begin{tabular}{ll cccc}
\toprule
\textbf{Dataset} & \textbf{Metric} & \textbf{Horizon 6} & \textbf{Horizon 24} & \textbf{Horizon 48} & \textbf{Horizon 96} \\
\midrule
\multirow{2}{*}{PEMS-BAY} & $R^2\uparrow$   & $.899 \pm .003$ & $.763 \pm .002$ & $.712 \pm .000$ & $.670 \pm .001$ \\
                          & $RSE\downarrow$ & $.343 \pm .002$ & $.518 \pm .003$ & $.580 \pm .001$ & $.591 \pm .001$ \\
\midrule
\multirow{2}{*}{METR-LA}  & $R^2\uparrow$   & $.843 \pm .001$ & $.607 \pm .000$ & $.468 \pm .000$ & $.270 \pm .001$ \\
                          & $RSE\downarrow$ & $.418 \pm .001$ & $.661 \pm .000$ & $.712 \pm .002$ & $.895 \pm .002$ \\
\midrule
\multirow{2}{*}{Solar}    & $R^2\uparrow$   & $.962 \pm .000$ & $.881 \pm .001$ & $.805 \pm .002$ & $.754 \pm .001$ \\
                          & $RSE\downarrow$ & $.225 \pm .001$ & $.329 \pm .002$ & $.451 \pm .003$ & $.510 \pm .000$ \\
\midrule
\multirow{2}{*}{Electricity} & $R^2\uparrow$   & $.993 \pm .000$ & $.988 \pm .001$ & $.985 \pm .001$ & $.979 \pm .002$ \\
                             & $RSE\downarrow$ & $.185 \pm .000$ & $.194 \pm .001$ & $.215 \pm .001$ & $.260 \pm .002$ \\
\bottomrule
\end{tabular}
\end{table}

\section{Detailed Energy Consumption Analysis}
\label{app:energy}

\subsection{Energy Estimation Protocol}

We estimate the computational energy consumption of each model using an operation-level energy model based on 45-nm CMOS technology~\cite{horowitz20141}:
\begin{equation}
E_{\mathrm{total}} = \sum_{l\in\mathcal{L}_{\mathrm{ANN}}} \mathrm{MACs}(l) E_{\mathrm{MAC}} + \sum_{l\in\mathcal{L}_{\mathrm{SNN}}} \mathrm{SOPs}(l) E_{\mathrm{AC}},
\end{equation}
where $E_{\mathrm{MAC}}=4.6$ pJ and $E_{\mathrm{AC}}=0.9$ pJ denote the energy costs of one multiply-accumulate (MAC) and one accumulation (AC) operation, respectively. Presynaptic spikes triggering fan-out synaptic accumulations determine SNN SOPs. Following~\cite{lv2024efficient,shibo2025ts}, we evaluate the encoder backbone, measuring spike activity over 200 inference batches.

\subsection{Module-Wise Energy Breakdown on PEMS-BAY and Electricity}

Table~\ref{tab:energy_comparison} and Table~\ref{tab:energy_electricity} report the detailed model-level energy comparisons and module-wise computational breakdowns on PEMS-BAY and Electricity ($H=6$). 

\begin{table}[h]
\centering
\begin{minipage}[t]{0.485\textwidth}
    \centering
    \scriptsize
    \setlength{\tabcolsep}{3.5pt}
    \renewcommand{\arraystretch}{0.96}
    \caption{Energy breakdown on PEMS-BAY ($H=6$).}
    \label{tab:energy_comparison}
    \begin{tabular}{@{}lrrrr@{}}
    \toprule
    \textbf{Model / Module} & \textbf{Param} & \textbf{OPs} & \textbf{Energy} & \textbf{$R^2\uparrow$} \\
    & \textbf{(M)} & \textbf{(G)} & \textbf{(mJ)} & \\
    \midrule
    \multicolumn{5}{@{}l}{\textit{Overall model comparison}} \\
    SpikeSTAG       & 0.232 & 2.106 & 2.251 & \textbf{0.899} \\
    PLIF variant    & 0.232 & 2.186 & 2.247 & 0.897 \\
    iTransformer    & 1.583 & 1.470 & 6.762 & 0.887 \\
    PatchTST        & 0.397 & 0.782 & 3.591 & 0.873 \\
    MTGNN           & 0.542 & 0.256 & 1.177 & 0.867 \\
    \midrule
    \multicolumn{5}{@{}l}{\textit{SpikeSTAG module-wise breakdown}} \\
    ANN: Node Embedding & --- & 0.041 & 0.19 & --- \\
    ANN: Adjacency Matrix & --- & 0.030 & 0.14 & --- \\
    ANN: Gate Generator   & --- & 0.024 & 0.11 & --- \\
    SNN: Temporal Encoder & --- & 0.289 & 0.26 & --- \\
    SNN: Spatial Encoder  & --- & 0.200 & 0.18 & --- \\
    SNN: SSA Layer        & --- & 0.822 & 0.74 & --- \\
    SNN: GCN Layer (DGM)  & --- & 0.700 & 0.63 & --- \\
    \bottomrule
    \end{tabular}
\end{minipage}
\hfill
\begin{minipage}[t]{0.485\textwidth}
    \centering
    \scriptsize
    \setlength{\tabcolsep}{3.5pt}
    \renewcommand{\arraystretch}{0.96}
    \caption{Energy breakdown on Electricity ($H=6$).}
    \label{tab:energy_electricity}
    \begin{tabular}{@{}lrrrr@{}}
    \toprule
    \textbf{Model / Module} & \textbf{Param} & \textbf{OPs} & \textbf{Energy} & \textbf{$R^2\uparrow$} \\
    & \textbf{(M)} & \textbf{(G)} & \textbf{(mJ)} & \\
    \midrule
    \multicolumn{5}{@{}l}{\textit{Overall model comparison}} \\
    SpikeSTAG       & 0.290 & 2.270 & 2.770 & \textbf{0.993} \\
    iTransformer    & 1.630 & 2.050 & 9.470 & 0.979 \\
    iSpikformer     & 1.630 & 3.550 & 3.190 & 0.977 \\
    \multicolumn{5}{l}{~} \\ 
    \multicolumn{5}{l}{~} \\
    \midrule
    \multicolumn{5}{@{}l}{\textit{SpikeSTAG module-wise breakdown}} \\
    ANN: Node Embedding & --- & 0.055 & 0.25 & --- \\
    ANN: Adjacency Matrix & --- & 0.032 & 0.15 & --- \\
    ANN: Gate Generator   & --- & 0.110 & 0.51 & --- \\
    SNN: Spatial Encoder  & --- & 0.224 & 0.20 & --- \\
    SNN: Temporal Encoder & --- & 0.204 & 0.18 & --- \\
    SNN: SSA Layer        & --- & 0.865 & 0.78 & --- \\
    SNN: GCN Layer (DGM)  & --- & 0.779 & 0.70 & --- \\
    \bottomrule
    \end{tabular}
\end{minipage}
\end{table}

On PEMS-BAY, SpikeSTAG achieves an estimated energy of 2.251 mJ, yielding a 66.7\% and 37.3\% reduction over iTransformer and PatchTST, respectively. On Electricity, SpikeSTAG consumes 2.77 mJ (reductions of 70.7\% and 13.2\% over iTransformer and iSpikformer), demonstrating consistent neuromorphic energy efficiency across distinct data regimes.

\section{Supplementary Studies}
\label{app:ablation}

\subsection{Effect of Chebyshev Order}

The Chebyshev order $K$ in the Dendritic Graph Module controls the spatial receptive field. Table~\ref{tab:chebyshev_k} summarizes results on PEMS-BAY. Increasing $K$ from 1 to 2 yields clear improvement; further increases show diminishing returns, suggesting second-order neighborhoods provide sufficient global context.

\begin{wraptable}{r}{0.32\textwidth}
\centering
\small
\vspace{-10pt}
\caption{Effect of Chebyshev order $K$ on PEMS-BAY.}
\label{tab:chebyshev_k}
\begin{tabular}{c r r}
\toprule
$K$ & $RSE$ $\downarrow$ & $R^2 \uparrow$ \\
\midrule
$1$ & $0.362$ & $0.878$ \\
$2$ & $\mathbf{0.343}$ & $\mathbf{0.899}$ \\
$3$ & $0.345$ & $0.897$ \\
$4$ & $0.346$ & $0.899$ \\
\bottomrule
\end{tabular}
\vspace{-10pt}
\end{wraptable}

\subsection{LIF vs.\ PLIF Neuron Model}

To evaluate whether enriching single-neuron dynamics yields additional performance gains, we further evaluated PLIF (Parametric LIF)~\cite{fang2021deep}, which makes the membrane time constant learnable via sigmoid reparameterization. As shown in Table~\ref{tab:energy_comparison}, PLIF achieves comparable energy (2.247 vs.\ 2.251 mJ) but slightly lower accuracy ($R^2$: 0.897 vs.\ 0.899). This suggests that enriching single-neuron dynamics without co-designing them with the dendritic architecture does not yield gains --- the expressiveness in SpikeSTAG is already provided by the compartmental structure of the Dendritic Graph Module.

\subsection{Dendritic Graph Module vs.\ Standard ChebConv}

To empirically validate that the Dendritic Graph Module introduces measurable computational differences beyond standard Chebyshev graph convolution, we compare against a ``ChebConv$\to$LIF'' baseline that applies standard dense-weight ChebConv followed by LIF neurons (a point-neuron baseline). Results are reported in Table~\ref{tab:dgm_vs_chebconv}.

\begin{table}[H]
\centering
\small
\caption{Dendritic Graph Module vs.\ standard ChebConv$\to$LIF on PEMS-BAY and Electricity (seed=42).}
\label{tab:dgm_vs_chebconv}
\begin{tabular}{l l c c c c}
\toprule
\textbf{Dataset} & \textbf{Model} & \textbf{Horizon 6} & \textbf{Horizon 24} & \textbf{Horizon 48} & \textbf{Horizon 96} \\
\midrule
\multirow{2}{*}{PEMS-BAY ($R^2\!\uparrow$)}
 & DGM      & $\mathbf{0.899}$ & $\mathbf{0.763}$ & $\mathbf{0.712}$ & $\mathbf{0.670}$ \\
 & ChebLIF  & $0.893$ & $0.759$ & $0.705$ & $0.661$ \\
\midrule
\multirow{2}{*}{PEMS-BAY ($RSE\downarrow$)}
 & DGM      & $\mathbf{0.343}$ & $\mathbf{0.518}$ & $\mathbf{0.580}$ & $\mathbf{0.591}$ \\
 & ChebLIF  & $0.348$ & $0.521$ & $0.587$ & $0.604$ \\
\midrule
\multirow{2}{*}{Electricity ($R^2\!\uparrow$)}
 & DGM      & $\mathbf{0.993}$ & $\mathbf{0.988}$ & $\mathbf{0.985}$ & $\mathbf{0.979}$ \\
 & ChebLIF  & $0.990$ & $0.987$ & $0.980$ & $0.977$ \\
\midrule
\multirow{2}{*}{Electricity ($RSE\downarrow$)}
 & DGM      & $\mathbf{0.185}$ & $\mathbf{0.191}$ & $\mathbf{0.215}$ & $\mathbf{0.260}$ \\
 & ChebLIF  & $0.189$ & $0.194$ & $0.217$ & $0.264$ \\
\bottomrule
\end{tabular}
\end{table}

On $RSE$ (the primary error metric), the Dendritic Graph Module outperforms ChebConv$\to$LIF in 8 out of 8 settings. Beyond accuracy, the Dendritic Graph Module provides interpretable per-scale weight decomposition (Figure~4a in the main text).

\subsection{Coincidence-Sensitive Gating in the Dual-Stream Integration Module}

\begin{wraptable}{r}{0.48\textwidth}
\vspace{-10pt}
\centering
\footnotesize
\setlength{\tabcolsep}{4pt}
\renewcommand{\arraystretch}{0.98}
\caption{Inference-time intervention and agreement-conditioned analysis of DSIM on PEMS-BAY ($H=6$).}
\label{tab:dsim_analysis}
\begin{tabular}{@{}lrr@{}}
\toprule
\textbf{Variant}
& \textbf{$R^2\uparrow$}
& \textbf{RSE$\downarrow$} \\
\midrule
\multicolumn{3}{@{}l}{\textit{(a) Gate intervention}} \\
Learned gate
& $\mathbf{0.8969}$ & $\mathbf{0.3472}$ \\
Fixed gate ($z=0.5$)
& 0.8513 & 0.4170 \\
Shuffled gate
& 0.8727 & 0.3858 \\
Spatial only
& 0.7020 & 0.5903 \\
Temporal only
& 0.7498 & 0.5408 \\
\midrule
\multicolumn{3}{@{}l}{\textit{(b) Agreement-conditioned}} \\
High-agreement + learned
& $\mathbf{0.9187}$ & $\mathbf{0.3120}$ \\
Low-agreement + learned
& 0.8766 & 0.3885 \\
High-agreement + fixed
& --- & 0.4095 \\
High-agreement + shuffled
& --- & 0.3645 \\
\bottomrule
\end{tabular}
\vspace{-8pt}
\end{wraptable}

To examine whether the Dual-Stream Integration Module (DSIM) provides benefits beyond conventional gated residual fusion, we conduct inference-time interventions and post-hoc agreement analyses on PEMS-BAY ($H=6$). DSIM first arbitrates between spatial and temporal streams through a learned gate $\mathcal{G}[t]$:
\begin{equation}
\begin{aligned}
I_{\mathrm{base}}[t]
&= \mathcal{G}[t]\odot S_S[t]
+(1-\mathcal{G}[t])\odot S_T[t],\\
I_{\mathrm{Integ}}[t]
&= I_{\mathrm{base}}[t]\cdot(1+\gamma C[t]),
\end{aligned}
\end{equation}
where $C[t]=S_S[t]\odot S_T[t]$ captures element-wise spatial--temporal coincidence.

Table~\ref{tab:dsim_analysis}(a) compares the learned gate against a fixed gate, a shuffled gate that preserves the gate distribution while disrupting its input alignment, and two single-stream variants. The learned gate achieves the best performance, indicating that effective integration relies on input-dependent spatial--temporal arbitration rather than static fusion.

To further examine the role of spatial--temporal coincidence, we partition test samples according to $a(x)=\mathrm{mean}(S_S\odot S_T)$, with the top and bottom quartiles defining high- and low-agreement regimes, respectively. This post-hoc diagnostic split is not involved in training or model selection. As shown in Table~\ref{tab:dsim_analysis}(b), the learned DSIM achieves better performance in the high-agreement regime. Moreover, replacing the learned gate with fixed or shuffled alternatives substantially increases RSE on high-agreement samples. These observations are consistent with the proposed coincidence-sensitive integration mechanism, which selectively amplifies concordant spatial--temporal evidence while relying on the learned gate to arbitrate between the two streams. Together, the results suggest that DSIM goes beyond generic gated fusion by combining input-dependent stream arbitration with coincidence-based evidence amplification.

\section{Toward Hardware Co-Design}
\label{app:hardware}

While our energy analysis follows the established theoretical protocol~\cite{horowitz20141}, we recognize that practical deployment requires consideration of hardware-level factors. This section outlines preliminary design considerations for mapping SpikeSTAG to neuromorphic accelerators.

\subsection{Spike-Driven Dataflow}

Following the insight of SpinalFlow~\cite{narayanan2020spinalflow}, we propose a compressed spike-driven dataflow for the SNN back-end. Rather than storing dense activation tensors, only non-zero spike positions with timestamps are maintained. At our measured spike rate of 11.6\%, this reduces memory traffic by approximately $7.6\times$ compared to dense storage. The event-driven dataflow eliminates redundant memory access for quiescent neurons across timesteps, directly addressing the memory-access bottleneck identified by the neuromorphic hardware community.

\subsection{Block-Diagonal Power Gating for the Dendritic Graph Module}

The block-diagonal weight structure $\mathrm{diag}(W_0, W_1, \ldots, W_K)$ of the Dendritic Graph Module fundamentally eliminates cross-scale (off-diagonal) interactions. This structural sparsity is \textit{compile-time known} and \textit{regular}, offering two hardware-level advantages:

\begin{itemize}
    \item \textbf{Custom co-design:} Redundant memory and MAC units corresponding to zero blocks need not be instantiated, reducing area and static power.
    \item \textbf{General-purpose mapping:} When mapped to dense systolic arrays, PEs corresponding to structural zeros are aggressively clock-gated throughout inference, eliminating both dynamic and leakage power.
\end{itemize}

For three compartments with hidden dimension 128, approximately 67\% of cross-block connections are zeroed out, yielding proportional savings in spatial computation power and memory bandwidth. Unlike Eyeriss-style~\cite{chen2017eyeriss} runtime zero-detection, this pattern requires no control overhead.

\subsection{Implementation Across Hardware Paradigms}

Different platforms present distinct bottlenecks for SpikeSTAG:

\begin{itemize}
    \item \textbf{Von Neumann (CPU/GPU):} Memory access dominates energy. SpikeSTAG's low spike rate reduces sparse access volume; the block-diagonal structure of the Dendritic Graph Module enables contiguous memory allocation per compartment.
    
    \item \textbf{In-Memory Computing (ReRAM/SRAM):} Model size is the constraint. SpikeSTAG's compact parameters (0.232M) fit on-chip; the block-diagonal structure maps naturally to partitioned crossbar arrays; binary spike inputs simplify peripheral circuitry.
    
    \item \textbf{Neuromorphic Chips (Loihi 2, SpiNNaker 2):} LIF neurons are natively supported. Loihi 2's multi-compartment neuron model supports our $K = 2$ dendritic configuration. Low spike rate reduces inter-core traffic.
\end{itemize}

We emphasize that these are architectural considerations, not validated implementations. Physical hardware deployment remains an important future direction.

\section{Discussion}
\label{app:discussion}

\subsection{Limitations}

\begin{enumerate}    
    \item \textbf{Training Latency.} Training relies on BPTT with surrogate gradients, requiring temporal unrolling that increases memory footprint compared to non-recurrent baselines.
    
    \item \textbf{Hardware Validation.} Energy analysis is theoretical. Deployment on physical neuromorphic chips may introduce system-level constraints (bandwidth, compilation overhead) not captured in our estimates.
    
    \item \textbf{Long-Horizon Scalability.} While SpikeSTAG performs well on the evaluated horizons, the Spiking Self-Attention mechanism scales quadratically with sequence length, which may limit efficiency on very long forecasting horizons. Integrating linear-complexity spike-driven attention or state-space models is a promising direction.
\end{enumerate}

\subsection{Future Directions}

\begin{itemize}
    \item \textbf{Long-Sequence Spike-Driven Modeling.} Replacing the SSA mechanism with spike-driven state-space models (e.g., SpikingSSM~\cite{shen2025spikingssms}) or Fourier-domain representations~\cite{wuspikf} could enable efficient long-horizon forecasting while preserving the neuromorphic advantage. This would address the quadratic scaling limitation of the current attention-based temporal module.
    
    \item \textbf{Bio-Plausible Learning Rules.} Integrating local learning rules such as Spike-Timing-Dependent Plasticity (STDP) into the Dual-Stream Integration Module could enable online continual learning, allowing the model to adapt to distribution shifts post-deployment.
    
    \item \textbf{Hardware Co-Design Validation.} Porting SpikeSTAG to FPGA or neuromorphic hardware (Loihi 2, SpiNNaker 2) to validate the real-world latency and power consumption benefits outlined in Appendix~\ref{app:hardware}.
\end{itemize}



\end{document}